\documentclass{bmvc2k}

\usepackage{epsfig}
\usepackage{xcolor}
\usepackage[linesnumbered,ruled,vlined]{algorithm2e}
\usepackage{tabularx}
\usepackage{booktabs}
\usepackage{enumitem}
\usepackage[misc]{ifsym}
\usepackage{tikz}
\usepackage{comment}
\usepackage{amsmath,amssymb} 
\usepackage{color}
\usepackage{graphicx}

\SetCommentSty{mycommfont}

\SetKwInput{KwInput}{Input}                
\SetKwInput{KwOutput}{Output}              

\DeclareMathOperator*{\argmax}{arg\,max}

\newcommand\blfootnote[1]{%
  \begingroup
  \renewcommand\thefootnote{}\footnote{#1}%
  \addtocounter{footnote}{-1}%
  \endgroup
}

\title{VL4Pose: Active Learning Through Out-Of-Distribution Detection For Pose Estimation}

\addauthor{Megh Shukla ${}^{\textrm{\Letter}}$ }{megh.shukla@epfl.ch}{1}
\addauthor{Roshan Roy ${}^{*}$}{roshan.roy@lmco.com}{3}
\addauthor{Pankaj Singh ${}^{*}$}{pankaj.singh@mercedes-benz.com}{2}
\addauthor{Shuaib Ahmed }{shuaib.ahmed@mercedes-benz.com}{2}
\addauthor{Alexandre Alahi }{alexandre.alahi@epfl.ch}{1}

\addinstitution{\small
 Visual Intelligence for Transportation Lab\\
 École Polytechnique Fédérale de Lausanne\\
 Lausanne, Switzerland
}
\addinstitution{
\small
 Mercedes-Benz Research and Development India\\
 Bengaluru, India
}
\addinstitution{
\small
 Lockheed Martin Corporation\\
 New Jersey, USA
}

\runninghead{Shukla \etal}{VL4Pose (Visual Likelihood for Pose Estimation)}

\def\etal{\emph{et al}\bmvaOneDot}

\begin{document}

\maketitle

\begin{abstract}
\noindent
Advances in computing have enabled widespread access to pose estimation, creating new sources of data streams. Unlike mock set-ups for data collection, tapping into these data streams through {\em on-device active learning} allows us to directly sample from the real world to improve the spread of the training distribution. However, on-device computing power is limited, implying that any candidate active learning algorithm should have a low compute footprint while also being reliable. Although multiple algorithms cater to pose estimation, they either use extensive compute to power state-of-the-art results or are not competitive in low-resource settings. We address this limitation with VL4Pose {\em (Visual Likelihood For Pose Estimation)}, a first principles approach for active learning through out-of-distribution detection. We begin with a simple premise: pose estimators often predict incoherent `poses' for out-of-distribution samples. Hence, can we identify a distribution of poses the model has been trained on, to identify incoherent poses the model is unsure of? Our solution involves modelling the pose through a simple parametric Bayesian network trained via maximum likelihood estimation. Therefore, poses incurring a low likelihood within our framework are out-of-distribution samples making them suitable candidates for annotation. We also observe two useful side-outcomes: VL4Pose in-principle yields better uncertainty estimates by unifying joint and pose level ambiguity, as well as the unintentional but welcome ability of VL4Pose to perform pose refinement in limited scenarios. We perform qualitative and quantitative experiments on three datasets: MPII, LSP and ICVL, spanning human and hand pose estimation. Finally, we note that VL4Pose is simple, computationally inexpensive and competitive, making it suitable for challenging tasks such as on-device active learning.

\blfootnote{\hspace*{-1.9em}${}^{*}$\textit{Pankaj and Roshan contributed equally. Work originated when Megh and Roshan were at Mercedes-Benz.}}
\end{abstract}

\section{Introduction}

\label{sec:intro}
Data-centric methods rely on developing models which are robust under a wide range of scenarios. However, training datasets are usually made in staged setups and may not capture the complexity associated with real world use cases. So how can we adapt our models to address different real world scenarios? One approach harnesses data streaming in from multiple end users, which represents the real world distribution. However, the resulting volume of data would overwhelm both, end user bandwidth and our data processing pipelines. Instead, can we turn to active learning \cite{settles2009active}, to sample only those images which the model considers informative?
\begin{figure}[t]
    \centering
    \includegraphics[width=\linewidth]{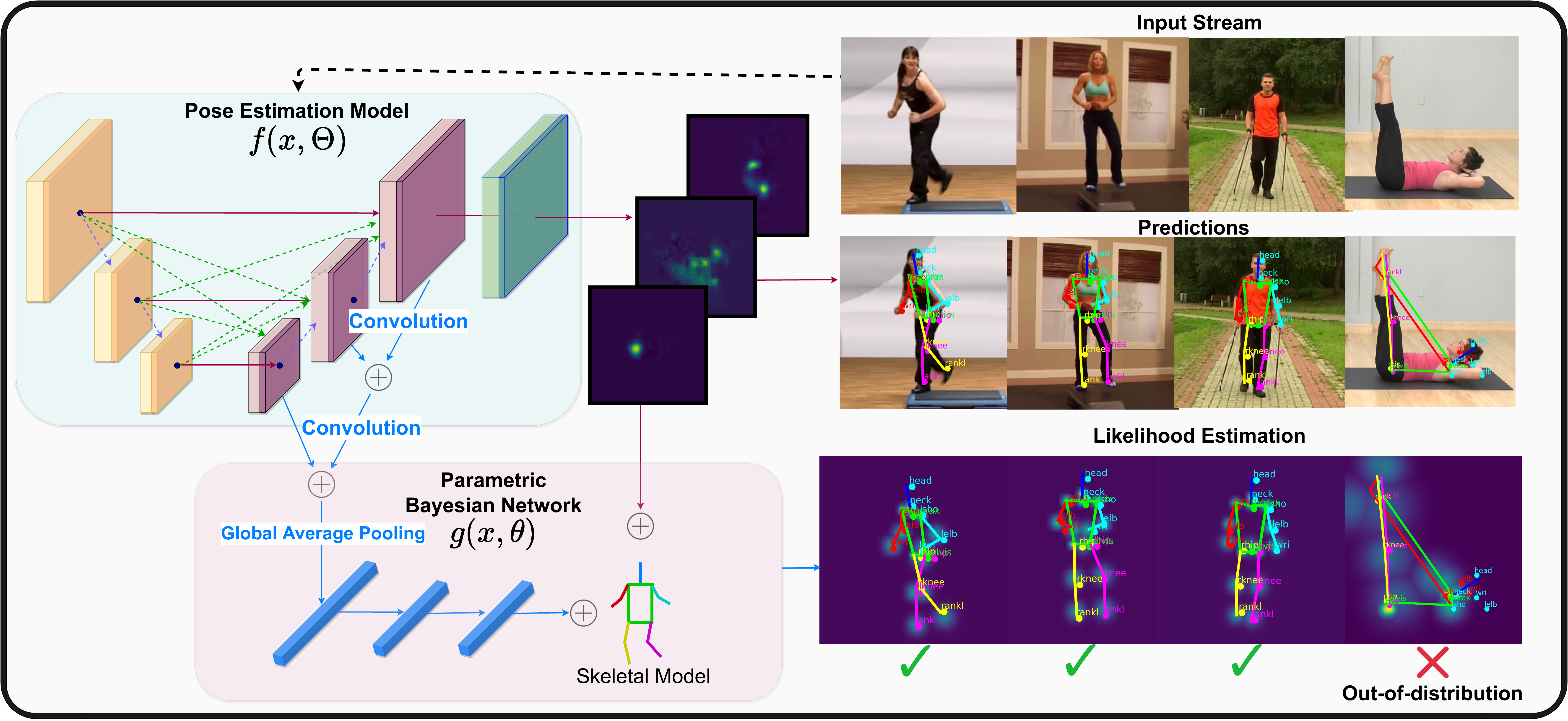}
    \caption{\textit{On-Device Active Learning}: Keeping on-device resource constraints in mind, we use a small (\textasciitilde 25\% size of the pose model) auxiliary network consisting of convolutional + fully connected layers to parameterize our skeletal model. The skeletal model is a simple Bayesian Network, trained via maximum likelihood to learn poses from the training distribution. Poses that incur a low likelihood are out-of-distribution and are added to the training dataset upon relabelling. VL4Pose delivers real time active learning at \textasciitilde 30FPS.}
    \label{fig:main}
\end{figure}

Active learning is a cyclical process of \textit{train $\Leftrightarrow$ sample and label new images} to improve the spread of the training distribution. Formally, the goal is to identify a subset of unlabelled data, which if labelled, imparts maximum information to the model thereby improving its performance. One could argue that the very essence of active learning is to perform out-of-distribution sampling. Indeed, out-of-distribution images will impart maximum information to a model trained on the training distribution. Active learning facilitates lower annotation costs as well as faster training and prototyping due to reduced data volumes. While active learning has been well studied in literature, keypoint estimation, and in particular human pose estimation presents a unique challenge. Specifically, popular human pose estimation architectures are fully convolutional and directly regress 2D heatmaps \cite{openpifpaf,zauss2021keypoint}. This prevents \cite{handpose,Shukla_2022_WACV} the use of many algorithms which rely on ensembles, dropouts or logits. As a result, a new wave of active learning methods \cite{liu2017active,Shukla_2022_WACV,learnloss,zhang2021deep, gong2022meta, handpose} address keypoint estimation. 

However, recent literature overlooks a key question: \textit{Where is active learning taking place?} The process can either be centralized in a computing cluster with no resource constraints, or decentralized on the end-user's device in a resource constrained environment. The latter, typically referred to as on-device active learning \cite{gudur2019activeharnet, qian2019distributed}, allows us to directly tap into the real world data distribution, while also opening new possibilities in customized machine learning for the end user. This is in contrast to centralized active learning which has indirect or no exposure to real time samples from the end user. \textit{While existing work focuses on pushing state-of-the-art in active learning for keypoint estimation, they impose high compute requirements which are infeasible for on-device learning.}

With \textit{VL4Pose}, we propose an algorithm for on-device active learning. We investigate a first principles approach to active learning for pose estimation; can we leverage simple pose constraints to identify out-of-distribution samples? Specifically, we model the skeletal structure through a Bayesian Network which captures simple conditional relationships between joints. These relationships are parameterized by a small auxiliary neural network which uses visual cues from images to maximize the likelihood of poses in our training data. Consequently, we expect that out-of-distribution images will have lower likelihood values, making them suitable candidates for annotation. We also show that our maximum likelihood formulation derives Multi-Peak Entropy \cite{liu2017active} and seamlessly unifies joint and pose level ambiguity, making it a better representative of uncertainty is comparison to \cite{ajain, handpose, kundu2022uncertainty}. Surprisingly, modelling simple skeletal constraints also facilitates pose refinement in limited scenarios, a first for an active learning algorithm. We validate our claims on two different tasks (human / hand pose) using two different architectures (direct keypoint regression / heatmaps) across three different datasets - MPII \cite{mpii}, LSP \cite{lsp,lspet} and ICVL \cite{Tang_2014_CVPR}. Our experiments show that VL4Pose has lower compute costs, interpretable and competitive with state-of-the-art, making it suitable for on-device real-time active learning.

\section{Related Work}

\noindent
\textbf{Active Learning}. Settles' survey \cite{settles2009active} is a comprehensive work on classical active learning covering algorithms based on diversity, ensemble and uncertainty sampling. The Query by Committee method of sampling (or ensembles) \cite{beluch2018power,korner2006multi,melville2004diverse} uses a family of hypothesis for active learning selection. Diversity based approaches include Core-Set \cite{sener2017active} which sequentially selects non-semantically similar points. Uncertainty estimation \cite{decomposition,gal2016dropout,gal2017deep,bacoun,kendall2017uncertainties,gp_heteroscedastic} provides quantitative measures to model ambiguity in the prediction. Bayesian Neural Networks have been traditionally used to estimate uncertainty; however recent approaches \cite{Shukla_2022_WACV,van2020uncertainty} explore computing uncertainty using a single forward pass through the network. Empirical approach to estimate ambiguity of the model include Learning Loss \cite{Shukla_2021_CVPR,learnloss} which similar to our approach uses an auxiliary neural network to predict the `loss' for an unlabelled image. Approaches that measure model change include expected gradient length \cite{settles2009active}, which uses the model's gradient as a directly proportionate measure of informativeness. Application domains \cite{egl_regression,huang2016active,settles2008analysis, Shukla_2022_WACV} of expected gradient length include image and text analysis. \\

\textbf{Pose Estimation}. Keypoint estimation (HPE) has been widely studied \cite{bulat2020toward, lauer2022multi, hg, hrnet, deeppose} in literature, with popular architectures regressing two dimensional heatmaps denoting the location of the joint. Heatmaps are preferred over direct keypoint regression since they retain positional accuracy which may otherwise have been lost with fully connected layers. While our work does not address multi-person pose estimation, our goal shares similarity with affinity fields \cite{paf,pifpaf,openpifpaf} used to associate parts with individual persons. Jain \textit{et al.} \cite{jain2013learning} and Tompson \textit{et al.} \cite{tompson2014joint} have previously leveraged Markov Random Fields to validate the configuration of poses predicted by the model. However, the priors defined by the approach do not scale well with rotations as well as changes in scale of the person. Moreover, belief propagation is expensive and takes a significant amount of time to converge. \textit{Unlike these methods, our goal is not to improve human pose estimation but instead detect out-of-distribution samples for active learning}. This allows us to use simpler models which converge faster and benefit from a low compute footprint. \\

\vspace{-3mm}

\textbf{Active Learning for Pose Estimation}. Challenges posed by popular human pose architectures (detailed in \cite{Shukla_2022_WACV, handpose}) have lead to the development of new algorithms. Liu and Ferrari \cite{liu2017active} proposed an intuition driven framework to model heatmap ambiguity. With VL4Pose, we provide a mathematical framework which not only incorporates heatmap ambiguity but also models spatial relationships between joints. Learning loss \cite{learnloss, Shukla_2021_CVPR} explored an idea parallel to out-of-distribution detection and identified which images were the most difficult for the model to learn. However, the method has limited ability in identifying out-of-distribution samples. The uncertainty modelled by EGL++ \cite{Shukla_2022_WACV} has high compute requirements. Recent state-of-the-art methods such as MATAL \cite{gong2022meta} and UncertainGCN \cite{Caramalau_2021_CVPR} use powerful techniques such as reinforcement learning and graph convolutional networks, however they are computationally very expensive. Uncertainty in 3D human pose \cite{wehrbein2021probabilistic,sengupta2021probabilistic, bertoni2019monoloco} utilizes depth information or stereo images, both of which are not available for general 2D pose estimation. We reserve our discussion on uncertainty in 2D for later.

\section{Methodology}

VL4Pose proposes active learning through out-of-distribution (OOD) for pose estimation. However, how do we define OOD in the context of pose estimation? For instance, \cite{hendrycks2016baseline, liang2017enhancing, ren2019likelihood} study OOD for classification, and limited literature addresses the same for pose estimation. Hence, VL4Pose frames OOD detection as a maximum likelihood problem; samples with a low likelihood in our framework have limited representation in the training set. Our premise is that pose estimators $f_{\Theta}$ may not generalize well beyond the training distribution. While the pose model accurately predicts the pose for images from the training set, more often than not the model makes errors on images from an unseen distribution. Formally, let $\hat{Y} = f (x, \Theta)$ where $\hat{Y} = \hat{y_1} \ldots \hat{y_N}$ represents the predicted joints and $x$ the input image. Can we identify images $x$ for annotation where $\hat{Y}$ is invalid?

\begin{figure}
    \centering
    \includegraphics[width=0.8\linewidth]{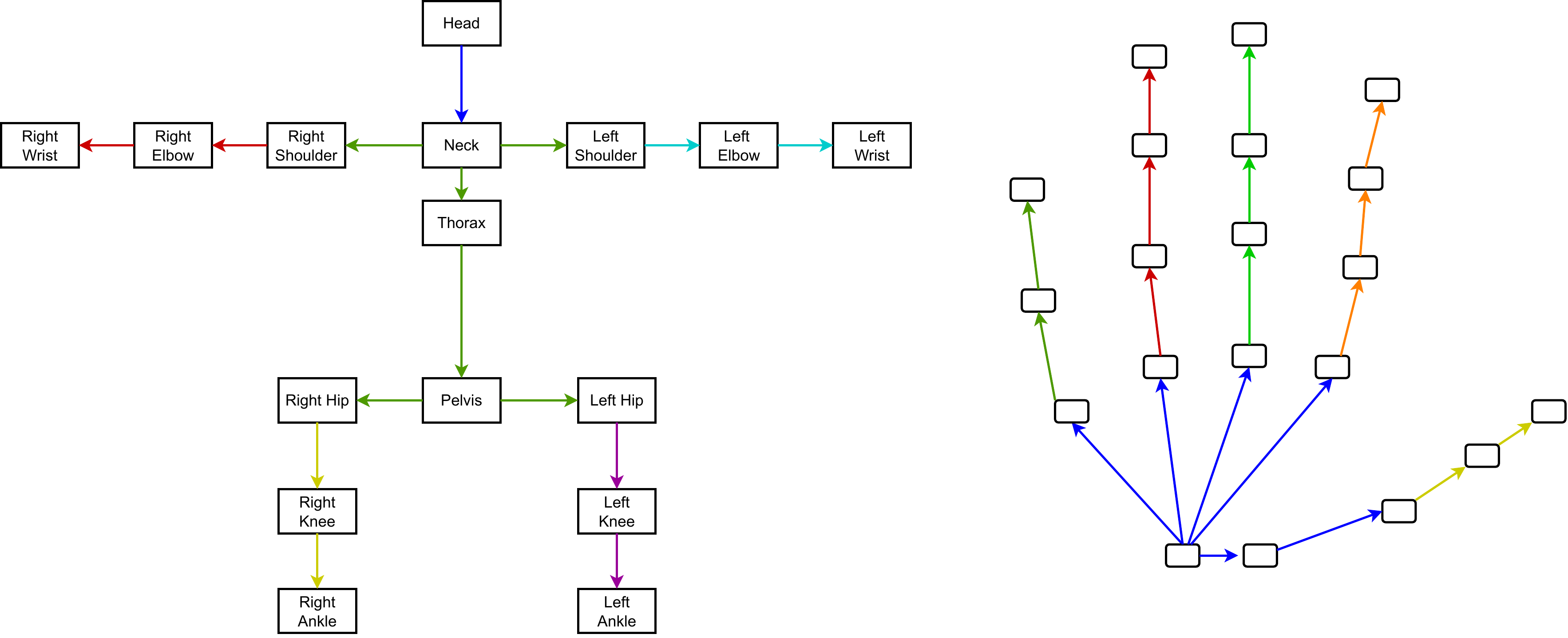}
    \caption{\textit{Skeletal Model}: While the pose estimator models joint localization $p_{pose}(Y=y)$, the parametric Bayesian network models the much simpler $q_{BN}(Y_1 = y_i | Y_2 = y_j)$ where the child is conditioned only on the parent joint. VL4Pose incorporates both $p_{pose}$ and $q_{BN}$ to model the distribution of poses for an image.}
    \label{fig:DAG}
\end{figure}

\subsection{Visual Likelihood Estimation}

Our solution lies in flipping the question: we learn to identify when $\hat{Y}$ is valid since the space of valid poses is much smaller than the set of invalid poses. However, what makes a pose valid? Indeed, a random collection of keypoints rarely makes a recognizable pose, the key to a valid pose is \textit{structure defined by the keypoints}. We represent this skeletal structure with a simple parametric Bayesian network (Fig. \ref{fig:DAG}, human and hand models), with singular parent-child chains. Such an approach allows us to exploit the Markov blanket principle of independence and simplify our analysis as well as model relations between various joints in an interpretable manner. Let $q(Y | x, \theta) = q(y_1, y_2 \ldots y_N | x, \theta)$ represents the distribution of joints $y_i$ and $\theta$ represents the parameters of the Bayesian network. Using the chain rule of probability and the Markov blanket, we can now decompose this distribution into a linear chain capturing pairwise dependencies:
\newpage
\vspace*{-3em}

\begin{align}
    q_{BN}(y_1, y_2 \ldots y_N | x, \theta) &= q(y_1 | y_2 \ldots y_N\, , x,  \theta ) q(y_2 |\,y_3 \ldots y_N, x, \theta ) \ldots q(y_N | x, \theta) \nonumber \\
    q_{BN}(y_1, y_2 \ldots y_N |x, \theta) &= \bigg[ \prod_{i=1}^{N-1} q(y_i | y_{i+1}, x, \theta) \bigg] q(y_N | x, \theta)
    \label{eq:chain}
\end{align}

Here $y_N$ represents the joint corresponding to the root node. We specifically note that this formulation is simple; for instance intuition suggests that the subset of joints forming the torso are not independent of each other, a fact that is oversimplified by our formulation. For instance, complex graphical models have been successfully used \cite{lehrmann2013non, wang2010two, park2004hierarchical, akhter2015pose} to improve pose estimation. \textit{However, we recall that our goal is to solve the simpler task of out-of-distribution detection}, and the proposed framework succeeds in achieving this objective.

Traditional application of maximum likelihood involves finding a set of parameters that maximizes the likelihood of our observations $X, Y$, where both X and Y are deterministic. However, the observation $X, Y$ need not be deterministic; for example in human pose estimation a joint can have multiple plausible locations based on the local maxima in the heatmap. Let $p_{pose}(Y) = p_{pose}(y_1, y_2 \ldots y_N) = \prod_{i=1}^{N} p(y_i)$ represent the pose estimator's (human/hand) distribution over the joints $y_1 \ldots y_N$. Note that our assumption of independence is in line with the training objective of popular pose estimators \cite{hg, hrnet, handpose}. The expected log-likelihood \textit{w.r.t} the set of keypoints is:
\begin{align}
     \mathbb{E}_{Y}\bigg[ \log\,\, q_{BN}(y_1, y_2 \ldots y_N | x, \theta) \bigg]
     \label{eq:expectation_1}
\end{align}
Substituting Eq. \ref{eq:chain} in Eq. \ref{eq:expectation_1} and expanding, we get (full derivation in supplementary):
\begin{equation}
    \sum_{Y}\bigg[ p_{pose}(y_N) \log\,\, q_{BN}(y_N | x, \theta)\, + \,  \sum_{i}^{N-1}\,p_{pose}(y_i) \log\, q_{BN}(y_i | y_{i+1}, X, \theta) \bigg]
    \label{eq:logll}
\end{equation}
Since $Y$ represents the set of keypoint random variables, expectation over $Y$ is the weighted likelihood over all possible pose configurations. Therefore, computing the expected likelihood allows us to incorporate a distribution over $Y$ into the framework. \textit{Intuitively, $p_{pose}$ allows us to model keypoint ambiguity whereas $q_{BN}$ models the the ambiguity associated with the entire pose.} The auxiliary network uses visual cues from the image to fit the parameters $\theta$ such that the likelihood over the training distribution is maximized. Therefore, poses incurring a low-likelihood correspond to out-of-distribution samples. Annotating such samples increases the spread of the training distribution resulting in better performing models. Since gradients of $\theta$ are detached from the pose estimator $\Theta$, we train the two networks simultaneously. During the training phase, we can directly observe both the parent $y_{i+1}$ as well as child joints $y_i$ to compute the likelihood. In the absence of ground truth (while performing active learning), we rely on the predictions of the pose estimator $\hat{Y} = f(x, \Theta)$ to estimate the likelihood.

The computation of $\sum_{Y}$ and $q_{BN}(y_i | y_{i+1}, X, \theta)$ depends upon the architecture (direct keypoint regression or heatmap) which we discuss now in greater detail.

\subsubsection{Direct Keypoint Regression}

\textbf{Modelling $q_{BN}(y_i | y_{i+1}, X, \theta)$}. We refer to the DeepPrior \cite{handpose} architecture, which uses fully connected layers to directly regress 3D keypoints ($y_i \in \mathbb{R}^{\textrm{joints} \times 3}$) for hand pose estimation. To impose conditional dependency, we predict the \textit{offset} to obtain the child joint given the parent joint: $y_i = y_{i+1} + \hat{o}_i$. The offset is learnt by the parametric network $\hat{o}_i = g_1 (x, \theta)$. Our belief is that the $y_i$ can be completely recovered given the parent $y_{i+1}$ and visual cues from image $\mathrm{X}$. Further, we also learn the covariance matrix $\Sigma_i = g_2(x, \theta)$ \cite{ajain, lu2022few, kendall2017uncertainties, dorta2018structured} that determines the spread around the offset for the child joint. Therefore:
\begin{equation}
    q_{BN}(y_i | y_{i+1}, x, \theta) = \mathcal{N} \bigg( y_{i} - [y_{i+1} + \hat{o}_i ],\, \Sigma_i \bigg)
\end{equation}

\noindent
\textbf{Modelling $\sum_{Y}$}. For hand pose estimation $Y =  f(x, \Theta)$ is a point estimate which implies there exists exactly one pose configuration; $p(y_i)$ = \{1 at ground truth location, 0 otherwise\} $\forall i$. This eliminates the need to sum over all possible pose configurations in Eq. \ref{eq:logll}, simplifying the computation.

\subsubsection{Heatmap Regression}

\textbf{Modelling $q_{BN}(y_i | y_{i+1}, X, \theta)$}. A significant portion of errors in human pose are due to incorrect association of left-right joints and keypoint swaps within the pose \cite{jiang2021regressive}. If we treat the offset as vectors, these sources of errors would cause some components of the offsets to average out during training leading to poor results. Hence we use an euclidean distance based measure which intuitively tries to find the optimal bone length $d_i = g(x, \theta)$ between the parent and child joint ($y_{i+1}$ and $y_i$ respectively). Specifically:
\begin{equation}
    q(y_i | y_{i+1}, x, \theta) = \mathcal{N} (\texttt{dist}(y_i, y_{i+1}) - \hat{d}_i\, , \, \sigma_i)
    \label{eq:distance}
\end{equation}
Learning the bone length is easier and is less strict in comparison to offsets, with our observations confirming that convergence is better for a distance based modelling approach for human pose estimation. \\

\vspace{-2mm}

\noindent
\textbf{Modelling $\sum_{Y}$}. During the training phase, we have access to the ground truth values which are point estimates for $Y =  f(x, \Theta)$. Therefore, the summation over all poses is reduced to one pose configuration to train the auxiliary network $\theta$ via maximum likelihood. However, during the active learning phase (no ground truth) we need to rely on the pose estimator's predictions which consists of heatmaps $h \in \mathbb{R}^{\textrm{joints} \times 64 \times 64}$ that represents a spatial probability distribution $p(y_i)$ for a joint $y_i \in \mathbb{R}^2$. Yet, computing the expectation over the entire heatmap is infeasible. Instead, we limit the domain of $p(y_i)$ to only the local maxima of the heatmap $h_i$ and thereby showing that Multi-Peak Entropy \cite{liu2017active} can be viewed as a consequence of maximizing the likelihood of poses. We then apply softmax normalization to provide a probabilistic interpretation of various local maxima being the true location of joint $i$. Therefore, we replace $p(y_i)$ with $\hat{p}(y_i) = \texttt{softmax}\,(\texttt{local\_maxima}\,(h_i))$ and the domain of $\hat{p}(y_i)$ now restricted to the location of the $\textrm{local\_maxima}$ of $h_i$.
\label{sec:heatmap}

\subsection{Discussion}

\textbf{Uncertainty For Pose Estimation}. Although multiple recent works \cite{ajain, liu2017active, handpose, kundu2022uncertainty, lu2022few} explore uncertainty for pose estimation, they suffer from a few limitations. Inspired by Kendall and Gal \cite{kendall2017uncertainties}, some approaches \cite{ajain, lu2022few, handpose} model heteroscedastic aleatoric uncertainty by learning a covariance matrix of the order $O(n^2)$ where $n$ represents the number of keypoints. This leads to two drawbacks: First, a much larger network is required to learn a fit for the covariance matrix. Second, larger networks tend to learn spurious correlations. Additionally, we believe that the definition of aleatoric uncertainty is misinterpreted in pose estimation (detailed in supplementary material) and hence we do not specify the uncertainty learnt by VL4Pose. Caramalau \textit{et al.} \cite{handpose} assumes independence between joints which is incorrect. Intuition suggests that observing a variable leads to a decrease in uncertainty for a correlated variable. While both Liu and Ferrari \cite{liu2017active} and Kundu \etal \cite{kundu2022uncertainty} share our approach of modelling joint level ambiguity, \cite{liu2017active} does not model pose uncertainty, whereas \cite{kundu2022uncertainty} uses a domain adaptation specific approach of modelling pose uncertainty. In contrast, VL4Pose provides a mathematical framework that incorporates both joint and pose level uncertainty for different network architectures. The method employs a linear chain of probabilities, therefore reducing the order of the variance / covariance matrix to $O(n)$. Further, VL4Pose directly captures the correlation between neighboring joints, limiting the possibility of learning spurious correlations. \\

\vspace{-2mm}

\textbf{Auxiliary Network}. The Bayesian network is parameterized through an auxiliary neural network as shown in Fig. \ref{fig:main}. Initially the network captures features from the pose estimator at various scales by using an appropriate convolutional kernel to downsize and add the larger feature map to the next smaller feature map. This is progressively done till we reach the smallest feature map, and subsequently perform global average pooling to obtain a one-dimensional feature. This is followed by simple fully connected layers, with the final layer predicting the parameters for each link in the skeleton. The network is trained to minimize the negative log-likelihood (Eq. \ref{eq:logll}). The two stage hourglass has $\approx 8.4M$ parameters, with the auxiliary network adding a further $\approx 2.1M$ parameters. \\

\vspace{-2mm}

\textbf{Hardware and Time Complexity}. We use two setups: Mobile computing (AMD Ryzen 5000, NVIDIA RTX 3060 Mobile) and server grade (Intel Xeon, NVIDIA V100). VL4Pose takes $\approx$ 30ms to process each image on both the setups (also implying that our hardware is not fully utilized). There are four real-time algorithms: VL4Pose, Learning Loss, Aleatoric uncertainty and Multi-Peak entropy. The former three have similar processing times ($\approx$ 30FPS) as they have near identical parametric networks, whereas Multi-Peak entropy is marginally faster since it benefits from a highly vectorized implementation. The inference time does not depend on the number of samples and have a complexity of $O(1)$ for a fixed model. In contrast, other algorithms are not designed for real-time use: graph based methods such as CoreSet, MCD-CKE, EGL++, GCN have a time complexity of at least $O(mn)$ since they model the interaction between all $m$ unlabelled and $n$ labelled samples. MATAL uses reinforcement learning which is compute intensive; the most efficient method takes 2 hours for active learning \cite{gong2022meta}. \\

\vspace{-2mm}

\begin{figure}[t]
    \centering
    \includegraphics[width=\linewidth]{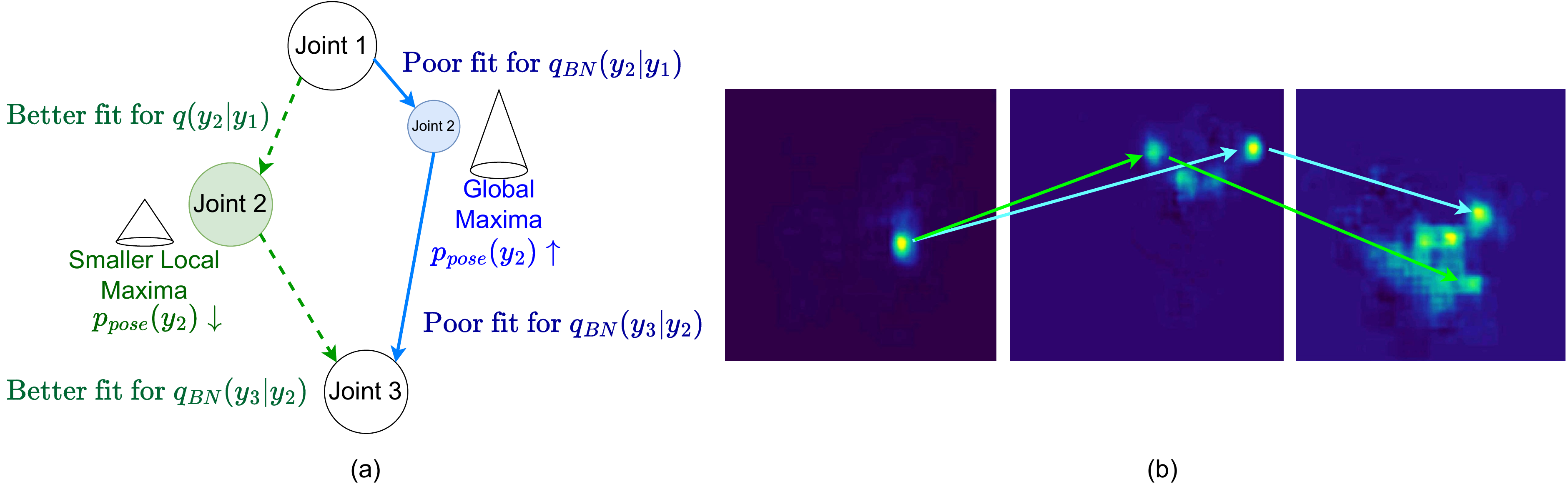}
    \caption{\textit{Pose Refinement}: Smaller but well placed local maxima are more likely to define a valid pose in comparison to poorly positioned global maxima.}
    \label{fig:refinement_concept}
\end{figure}

\textbf{Pose Refinement: Human Pose Estimation}. Surprisingly, a minor tweak to VL4Pose can also allow us to perform pose refinement \cite{moon2019posefix, nie2018hierarchical, wang2020graph, kamel2020hybrid} in certain scenarios, which to the best of our knowledge is the first active learning algorithm to do so. We categorically state that we do not intend to compete with state-of-the-art in pose refinement, but our intention is to highlight the versatility associated with VL4Pose in comparison to other active learning algorithms. Fig. \ref{fig:refinement_concept} provides some intuition into the interplay between skeletal structure $q_{BN}$ and heatmap ambiguity $p_{pose}$. Conventional human pose estimation approaches rely on inferring keypoints as the global maxima of their respective heatmaps. \textit{However, certain poses may have a higher likelihood by incorporating local maxima that better explain the skeletal structure.} Mathematically, instead of computing the expected likelihood over all poses (Eq. \ref{eq:logll}), we find the pose configuration that results in the highest likelihood: 
\begin{equation}
    Y^* = \argmax \bigg[ p_{pose}(y_N) \log\,\, q_{BN}(y_N | x, \theta)\, + \,  \sum_{i}^{N-1}\,p_{pose}(y_i) \log\, q_{BN}(y_i | y_{i+1}, X, \theta) \bigg]    
\end{equation}

\section{Experiments}

\begin{figure}[t!]
\centering     
\subfigure
{
    \label{fig:likelihood_plots}
    \includegraphics[width=0.95\linewidth]{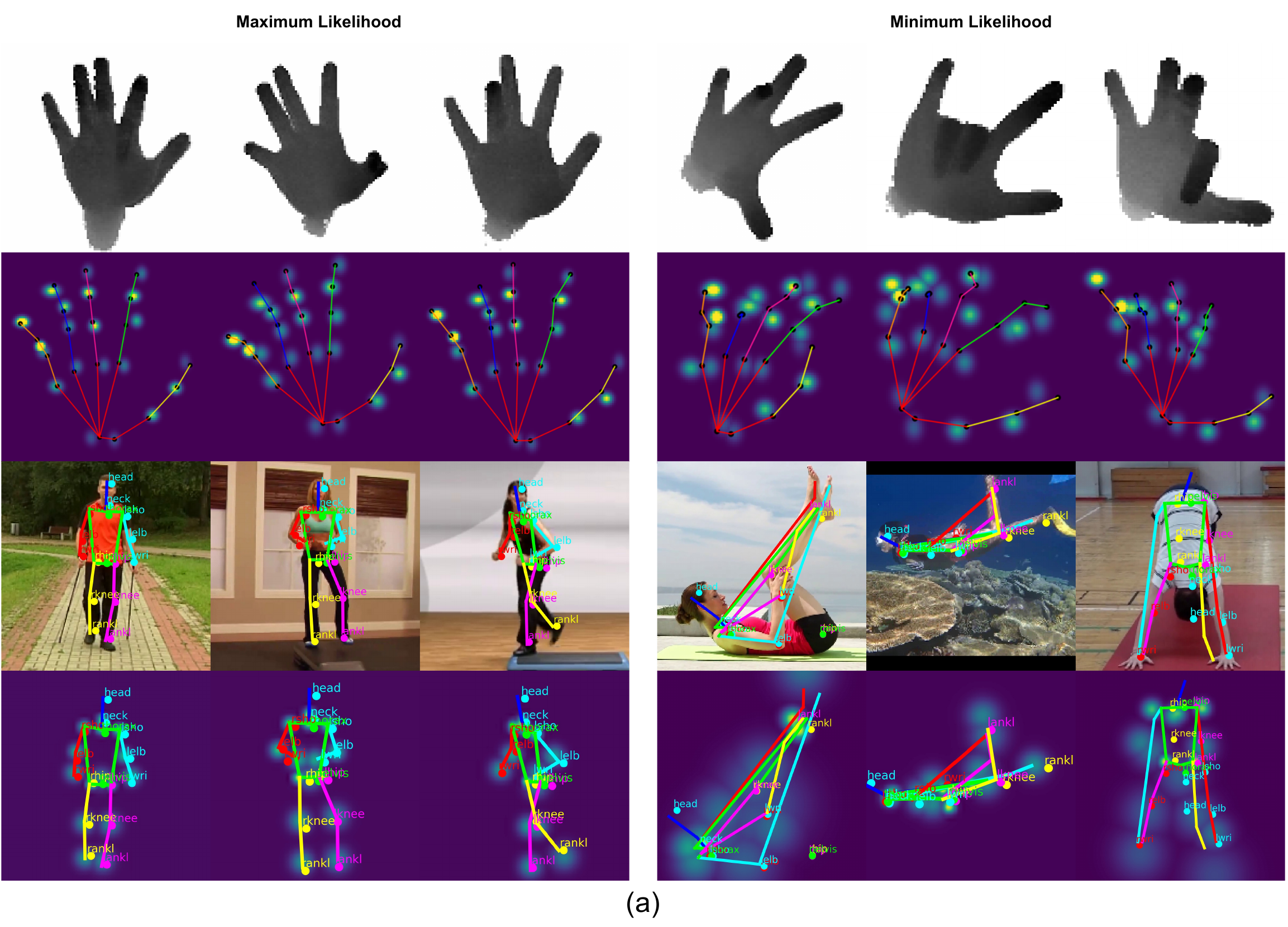}
}
\subfigure
{
    
    \label{fig:refinement_plots}
    \includegraphics[width=0.95\linewidth]{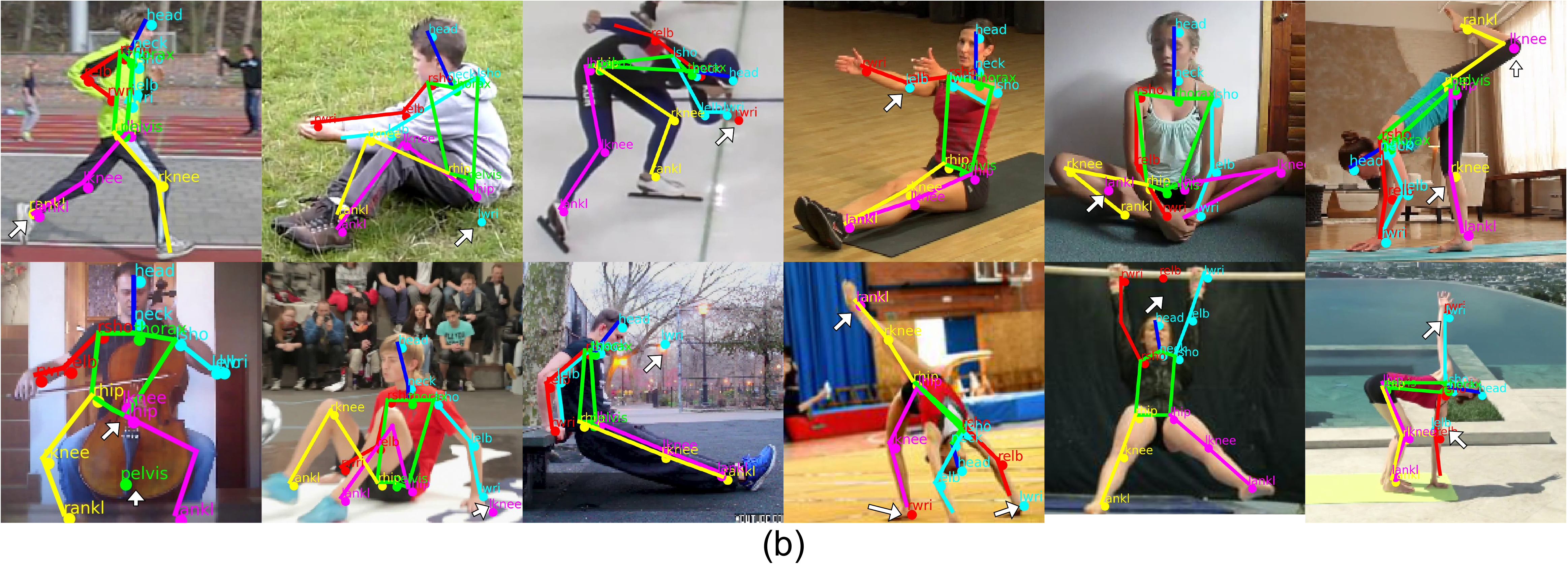}
}
\caption{[\textit{Please zoom in}] (a) \textit{Visualizing} $q_{BN}(y_i | y_{i+1}, x, \theta)$: The skeleton represents the pose estimator's predictions $\hat{Y} = f(x, \Theta)$ and filled circles are the ground truth $Y$. We highlight the correlation between pose uncertainty and likelihood, and likelihood with actual model performance (b) \textit{Pose refinement}: The skeleton represents the optimal pose configuration $Y^*$ that maximizes the likelihood, and filled circles are the the pose estimator's predictions $\hat{Y} = f(x, \Theta)$. We highlight VL4Pose's potential to identify the correct pose $Y^*$ even when $\hat{Y}$ has \underline{minor} errors (marked in arrows). \textit{Additional images in supplementary.}}
\end{figure}

\textbf{Goal}: VL4Pose is an algorithm for \textit{on-device active learning}, which is not only competitive but also works in real-time due to lower compute costs. Our experiments cover uncertainty and likelihood estimation, which forms the core of VL4Pose. We provide qualitative visualizations as well as quantitative comparisons through active learning to highlight how the uncertainties learnt by the model play a role in out-of-distribution detection. Our code is available at: \url{https://github.com/meghshukla/ActiveLearningForHumanPose}  

\textbf{Qualitative analysis}. Fig. \ref{fig:likelihood_plots} is generated by visualizing the distributions corresponding to \textit{offsets} for hand pose and \textit{distances} for human pose (visualization method described in the supplementary material). We observe that images with a low log-likelihood are a case where the joint predictions do not match the conditional distribution and are out-of-distribution. We observe that the accuracy of model predictions is highly correlated with the log-likelihood value; poor scores are correlated to poor predictions. Another important observation is that highly correlated parent-child joints have a higher degree of certainty given the parent. Previous methods discarded correlation, and hence double counted the uncertainty for highly correlated joints. Fig. \ref{fig:refinement_plots} highlights VL4Pose's ability to refine poses when the model gets a few joints wrong. VL4Pose is the first active learning algorithm to perform pose refinement since pose is explicitly being modelled by our method.

\textbf{Quantitative analysis}. We use the same active learning experiment setup as in \cite{Shukla_2021_CVPR,Shukla_2022_WACV,learnloss, handpose}. We conduct our experiments on human pose using two single person datasets: MPII \cite{mpii} and LSP/LSPET \cite{lsp,lspet}. For hand pose, we use the ICVL dataset \cite{Tang_2014_CVPR}. Following \cite{liu2017active} we subsample MPII to contain images with persons having all joints. Since VL4Pose does not model diversity, we follow \cite{learnloss,handpose} by performing an initial round of random sampling every active learning cycle followed by VL4Pose (for ICVL dataset only). We differ from \cite{Shukla_2022_WACV} in two aspects: we train the network to predict occluded joints and take larger crops around persons to add variability. We use different model architectures (Stacked Hourglass [Human Pose]\cite{hg} - following \cite{learnloss, Shukla_2022_WACV} and DeepPrior [Hand Pose] - defined and followed by \cite{Caramalau_2021_CVPR, handpose}). Tab. \ref{tab:res} shows that VL4Pose performs favourably amongst all algorithms with low compute (capable of running at 30FPS): Learning Loss \cite{learnloss}, Aleatoric Uncertainty \cite{handpose}, Multi-peak entropy \cite{liu2017active}. Multi-Peak is architecture specific and does not extend to hand pose. Learning Loss and Aleatoric lack in performance since they do not explicitly model the pose. In comparison to state-of-the-art (MATAL \cite{gong2022meta} and GCN \cite{Caramalau_2021_CVPR}), VL4Pose reports competitive but slightly inferior numbers. However, both methods are compute intensive and cannot be used for real-time on-device active learning.

\section{Conclusion} 

VL4Pose is about making the small things click; how far can incorporating simple domain knowledge take us? We answer this with a framework that models simple skeletal constraints to identify out-of-distribution samples. The method seamlessly unifies joint and pose level uncertainty, allowing for better uncertainty estimates. We qualitatively and quantitatively assess VL4Pose, noting that the method is interpretable, real-time, architecture agnostic and competitive with the state-of-the-art, making it suitable for on-device active learning. While pose refinement was an unintended consequence, we lay foundation for future work to incorporate complex skeletal models to push the barriers for out-of-distribution, active learning and pose refinement simultaneously.

\begin{table}[t]
\centering
\resizebox{0.6\columnwidth}{!}
{%
\renewcommand{\arraystretch}{0.7}
\begin{tabular}{ll@{\hskip 0.1in}l@{\hskip 0.1in}|@{\hskip 0.1in}l@{\hskip 0.1in}l@{\hskip 0.1in}|@{\hskip 0.1in}l@{\hskip 0.1in}l@{\hskip 0.1in}|@{\hskip 0.1in}l@{\hskip 0.1in}l@{\hskip 0.1in}|@{\hskip 0.1in}l@{\hskip 0.1in}l@{\hskip 0.1in}}

\multicolumn{11}{c}{\textbf{MPII: (PCKh, Percentage Correct Keypoints - head) Mean}$\pm$ \textbf{Std. Dev. (5 runs)}}\\

\toprule
\#images $\rightarrow$  &  \multicolumn{2}{|c|}{2000} & \multicolumn{2}{c|}{3000} & \multicolumn{2}{c|}{4000} & \multicolumn{2}{c|}{5000} & \multicolumn{2}{c}{6000}\\
\midrule
Methods & $\mu$ & $\sigma$ & $\mu$ & $\sigma$ & $\mu$ & $\sigma$ & $\mu$ & $\sigma$ & $\mu$ & $\sigma$ \\ 

\midrule
Random & 78.05 & 0.55 & 81.59 & 0.33 & 84.2 & 0.27 & 85.94 & 0.23 & 87.43 & 0.21 \\

Core-set \cite{sener2017active} & 75.66 & 1.22 & 80.52 & 1.15 & 84.1 & 0.54 & 86.08 & 0.35 & 87.68 & 0.31 \\

Learning Loss \cite{learnloss, Shukla_2021_CVPR} & 77.09 & 0.96 & 82.47 & 0.46 & 85.15 & 0.36 & 86.83 & 0.28 & 87.95 & 0.24 \\

EGL++ \cite{Shukla_2022_WACV} & 78.54 & 0.71 & 82.03 & 0.61 & 84.39 & 0.22 & 86.11 & 0.32 & 87.78 & 0.48 \\

Aleatoric \cite{kendall2017uncertainties} & 76.07 & 1.12 & 82.03 & 0.52 & 84.62 & 0.62 & 86.76 & 0.62 & 88.0 & 0.25  \\

Multi-peak \cite{liu2017active} & 82.49 & 0.72 & 84.62 & 0.46 & 85.88 & 0.31 & 87.47 & 0.51 & 88.54 & 0.71 \\

\textbf{VL4Pose} & 82.3 & 0.93 & 84.71 & 0.72 & 86.16 & 0.66 & 87.71 & 0.33 & 88.96 & 0.38 \\
\bottomrule
%
\end{tabular}
}
\vspace{2mm}

\resizebox{0.6\columnwidth}{!}{%
\renewcommand{\arraystretch}{0.7}
\begin{tabular}{ll@{\hskip 0.1in}l@{\hskip 0.1in}|@{\hskip 0.1in}l@{\hskip 0.1in}l@{\hskip 0.1in}|@{\hskip 0.1in}l@{\hskip 0.1in}l@{\hskip 0.1in}|@{\hskip 0.1in}l@{\hskip 0.1in}l@{\hskip 0.1in}|@{\hskip 0.1in}l@{\hskip 0.1in}l@{\hskip 0.1in}}

\multicolumn{11}{c}{\textbf{LSP and LSPET: Mean} $\pm$ \textbf{Sigma (5 runs)}}\\

\toprule
\#images $\rightarrow$  &  \multicolumn{2}{|c|}{2000} & \multicolumn{2}{c|}{3000} & \multicolumn{2}{c|}{4000} & \multicolumn{2}{c|}{5000} & \multicolumn{2}{c}{6000}\\
\midrule
Methods & $\mu$ & $\sigma$ & $\mu$ & $\sigma$ & $\mu$ & $\sigma$ & $\mu$ & $\sigma$ & $\mu$ & $\sigma$ \\ 

\midrule
Random & 74.24 & 0.68 & 76.91 & 0.91 & 79.11 & 0.64 & 80.56 & 0.35 & 81.47 & 0.65 \\

Core-set \cite{sener2017active} & 74.26 & 0.61 & 76.89 & 0.96 & 79.06 & 0.39 & 80.14 & 0.47 & 80.94 & 0.50 \\

Learning Loss \cite{learnloss, Shukla_2021_CVPR} & 73.99 & 0.28 & 76.71 & 0.63 & 78.53 & 0.37 & 79.91 & 0.37 & 80.77 & 0.24 \\

EGL++ \cite{Shukla_2022_WACV} & 74.51 & 1.02 & 77.32 & 0.69 & 79.26 & 0.69 & 80.68 & 0.45 & 81.76 & 0.24 \\

Aleatoric \cite{kendall2017uncertainties} & 74.24 & 0.60 & 76.94 & 0.47 & 79.15 & 0.62 & 80.11 & 0.40 & 80.91 & 0.49 \\

Multi-peak \cite{liu2017active} & 77.24 & 0.61 & 79.56 & 0.46 & 81.29 & 0.31 & 82.81 & 0.5 & 83.11 & 0.71 \\

\textbf{VL4Pose} & 77.36 & 0.68 & 79.71 & 0.50 & 81.48 & 0.46 & 82.75 & 0.49 & 83.69 & 0.47 \\

\bottomrule
\end{tabular}
}
\vspace{2mm}

\resizebox{0.6\columnwidth}{!}{%
\renewcommand{\arraystretch}{0.7}
\begin{tabular}{ll@{\hskip 0.1in}l@{\hskip 0.1in}|@{\hskip 0.1in}l@{\hskip 0.1in}l@{\hskip 0.1in}|@{\hskip 0.1in}l@{\hskip 0.1in}l@{\hskip 0.1in}|@{\hskip 0.1in}l@{\hskip 0.1in}l@{\hskip 0.1in}|@{\hskip 0.1in}l@{\hskip 0.1in}l@{\hskip 0.1in}}

\multicolumn{11}{c}{\textbf{ICVL: (MSE, Mean Square Error) Mean} $\pm$ \textbf{Std. Dev. (5 runs)}}\\

\toprule
\#images $\rightarrow$  &  \multicolumn{2}{|c|}{200} & \multicolumn{2}{c|}{400} & \multicolumn{2}{c|}{600} & \multicolumn{2}{c|}{800} & \multicolumn{2}{c}{1000}\\
\midrule
Methods & $\mu$ & $\sigma$ & $\mu$ & $\sigma$ & $\mu$ & $\sigma$ & $\mu$ & $\sigma$ & $\mu$ & $\sigma$ \\ 

\midrule

Random & 16.83 & 0.84 & 15.31 & 0.45 & 14.12 & 0.53 & 13.68 & 0.44 & 13.21 & 0.35 \\
Core-set \cite{sener2017active} & 16.86 & 0.74 & 14.73 & 0.46 & 14.02 & 0.81 & 13.69 & 0.49 & 13.43 & 0.44 \\
MCD-CKE \cite{handpose} & 19.88 & 0.38 & 15.06 & 0.43 & 13.61 & 0.46 & 12.85 & 0.71 & 12.54 & 0.59 \\
CoreGCN \cite{Caramalau_2021_CVPR} & 17.84 & 0.44 & 14.68 & 0.53 & 13.27 & 0.56 & 12.91 & 0.81 & 12.69 & 0.41 \\

\textbf{VL4Pose} & 16.87 & 0.47 & 14.89 & 0.52 & 13.64 & 0.51 & 13.02 & 0.77 & 12.84 & 0.54 \\

\bottomrule
\multicolumn{11}{c}{\Large } \\
\end{tabular}
}
\caption{\textit{Active Learning Simulation}: Human Pose: MPII, LSP-LSPET and Hand Pose: ICVL. Both PCK/PCKh and Mean Square Error (MSE) indicate accuracy of predictions, with higher values being better for PCK/PCKh and lower the better for MSE.}
\label{tab:res}
\end{table}

\newpage
\bibliography{egbib}
\end{document}


\maketitle

\begin{abstract}

\noindent
This supplementary material covers the following: 1) Deriving our expected likelihood formulation 2) Short discussion of aleatoric uncertainty in pose estimation 3) Description of the method to visualize our conditional distribution 4) Additional images for likelihood estimation and pose refinement 5) Algorithm implementation: VL4Pose

\end{abstract}

\section{Likelihood Estimation}

Our skeleton formulation allows us to describe the distribution over joints as: 
\begin{align}
    q_{BN}(y_1, y_2 \ldots y_N |x, \theta) &= \bigg[ \prod_{i=1}^{N-1} q(y_i | y_{i+1}, x, \theta) \bigg] q(y_N | x, \theta)
    \label{eq:chain}
\end{align}
Here `N' represents the number of joints with $q(y_N | x, \theta)$ representing the distribution of the root node, such as the head joint. We have also defined the pose estimator's distribution over the joints:
\begin{equation}
    p_{pose}(Y) = p_{pose}(y_1, y_2 \ldots y_N) = \prod_{i=1}^{N} p(y_i)
\end{equation}
As previously noted, our assumption of independence is in line with the training objective of popular pose estimators \cite{hg, hrnet, handpose}. We also note that $Y$ denotes the set of random variables $y_1 \ldots y_N$. The expected log-likelihood \textit{w.r.t} the set of keypoints is:
\begin{align}
     \mathbb{E}_{Y}\bigg[ \texttt{log}\,\, q_{BN}(y_1, y_2 \ldots y_N | x, \theta) \bigg]
     \label{eq:expectation_1}
\end{align}
Substituting Eq: \ref{eq:chain} in Eq: \ref{eq:expectation_1} and expanding, we get:

\begin{equation}
    \mathbb{E}_{Y}\bigg[\texttt{log}\,\, q_{BN}(y_N | x, \theta)\, + \,  \sum_{i}^{N-1}\, \texttt{log}\, q_{BN}(y_i | y_{i+1}, X, \theta) \bigg]
    \label{eq:logll}
\end{equation}

For human pose, the domain of $p(Y)$ represents all possible positions in the heatmap across all joints, which is intractable to compute. Hence, we limit the domain to local maxima in the heatmap for all joints. To ensure that the resultant distribution is valid, we normalize the local maxima within a heatmap to sum upto one. For hand pose, the network does not predict a distribution but provides a point estimate of $y$, limiting the domain to one pose configuration only. Therefore we can represent Eq: \ref{eq:logll} as:
\begin{equation}
    \sum_{Y}\bigg[\texttt{log}\,\, q_{BN}(y_N | x, \theta)\,\prod_{i=1}^{N} p(y_i)\, + \,  \sum_{i}^{N-1}\, \texttt{log}\, q_{BN}(y_i | y_{i+1}, X, \theta)\,\prod_{i=1}^{N} p(y_i)\, \bigg]
\end{equation}
We make one important note: $q(y_i | \ldots)$ depends only on $p(y_i)$, therefore $\sum_Y \prod_{j=1}^{N} p(y_j) = 1$ where $j \neq i$. Hence, we arrive at our final formulation:
\begin{equation}
    \sum_{Y}\bigg[ p_{pose}(y_N) \texttt{log}\,\, q_{BN}(y_N | x, \theta)\, + \,  \sum_{i}^{N-1}\,p_{pose}(y_i) \texttt{log}\, q_{BN}(y_i | y_{i+1}, X, \theta) \bigg]
    \label{eq:final}
\end{equation}



\section{Aleatoric Uncertainty For Pose Estimation}

Since our approach is similar to that in \cite{ajain,handpose, lu2022few}, one might be tempted to brand this method as aleatoric uncertainty. However, we consciously refrain from doing so. Aleatoric uncertainty represents the noise inherent in our data which cannot be reduced by increasing the samples drawn. While \cite{ajain, lu2022few} and \cite{handpose} further define aleatoric uncertainty for human and hand pose respectively, we believe that aleatoric uncertainty for keypoints is incorrectly represented in this literature.

Caramalau \textit{et al.} \cite{handpose} directly extends \cite{kendall2017uncertainties} to hand keypoint estimation, thereby solving $p(\texttt{joints} | X, \Theta) = \prod_i p(y_i | X, \theta)$. The assumption that all joints are independent of each other is incorrect when computing aleatoric uncertainty. Observing any one of these variable results in drastic uncertainty reduction for the unobserved counterpart, going against the principle of aleatoric uncertainty.

Works such as \cite{ajain, lu2022few} model aleatoric uncertainty as a multivariate normal distribution over joints $Y$. However, this uncertainty is reducible by observing more data and thus not aleatoric in a strict sense. For instance, rare poses are difficult to learn and hence the any network estimates a covariance matrix that reflects the uncertainty in the model's predictions. However, if we sample and train the model on more of such rare poses, it is expected that the model performance improves on these poses thereby reducing the associated uncertainty. This is in conflict with the definition of aleatoric uncertainty. Hence, we refrain from categorizing our uncertainty measures as aleatoric or epistemic since more investigation is required into sources of uncertainty for human pose.

\section{Visualizing the conditional distribution}

Visualizing the offset based conditional distribution for hand pose estimation is trivial. The normal distribution for the child joint is centred around the point determined by the parent joint adjusted with the predicted offset. Instead of visualizing in 3D, we visualize the distribution in 2D which is better suited for print media. This requires marginalizing over the depth $d$ since we wish to preserve the spatial representation for the multivariate normal distribution. Fortunately for us, marginalizing over a multivariate normal distribution is equivalent to dropping the variable being marginalized from the mean and covariance matrix of the distribution. Therefore, the resultant spatial normal distribution obtained by marginalizing the depth is straightforward to visualize in 2D.

In contrast, visualizing the distance based conditional distribution for human pose estimation is tricky. Viewing the distribution as a ring for all the skeletal links with radius, thickness as per the predicted mean, variance soon results in an overlapping non-informative visualization. Instead, for each link we plot a univariate gaussian in 2D with its centre located at the predicted distance along the line joining the parent and the child. Our reasoning for following this approach is based on the triangle inequality, where the difference between the predicted and actual distance is the lowest when the predicted gaussian lies along the same axis as the parent-child joints. Fortunately, this approach is easier for the visualization and analysis of multiple joints simultaneously as shown in the paper.

\section{Algorithm Implementation: VL4Pose}

Algorithm : \ref{algo:1} provides a pseudo-code for implementing VL4Pose. The essence of the pseudo-code lies in depth first search to evaluate the likelihood for various poses. The directed acyclic graph is represented as a tree with each node representing the joint. Each joint is associated with peaks and locations (obtained from joint heatmap) as well as parameters associated with the parent-child distribution. The pseudo-code recursively evaluates the combination of joints which results in the highest expected likelihood. The pseudo-code can be easily tweaked to obtain the highest likelihood as well as pose from the resultant heatmaps and conditional distributions.

\section{Visualizations}

\begin{figure}
    \centering
    \includegraphics[width=\linewidth]{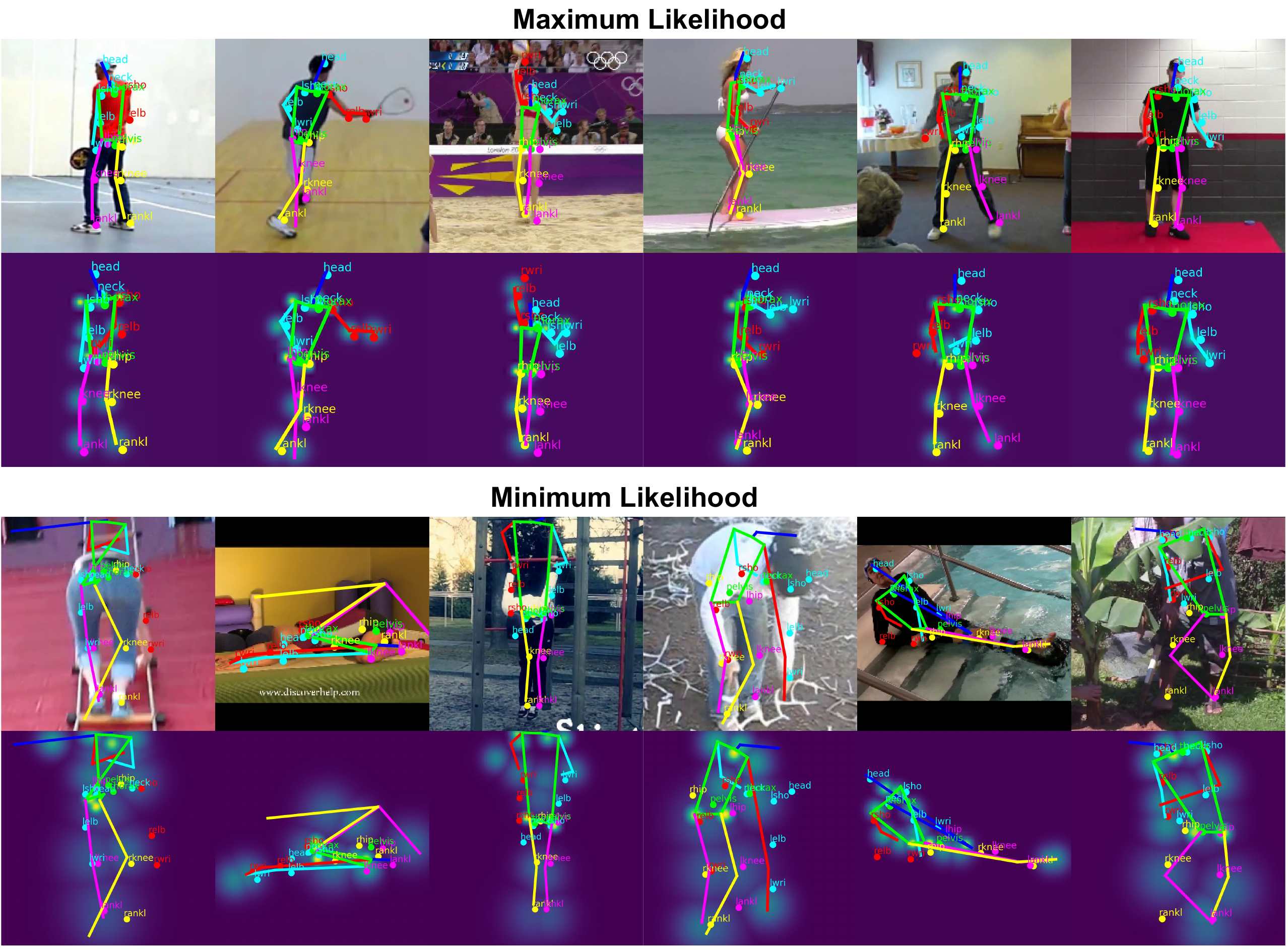}
    \caption{[Please zoom in] \textit{Visualizing} $q_{BN}(y_i | y_{i+1}, x, \theta)$: The skeleton represents the pose estimator's predictions $\hat{Y} = f(x, \Theta)$ and filled circles are the ground truth $Y$. We highlight the correlation between pose uncertainty and likelihood, and likelihood with actual model performance.}
    \label{fig:supp_pose}
\end{figure}

\begin{figure}
    \centering
    \includegraphics[width=\linewidth]{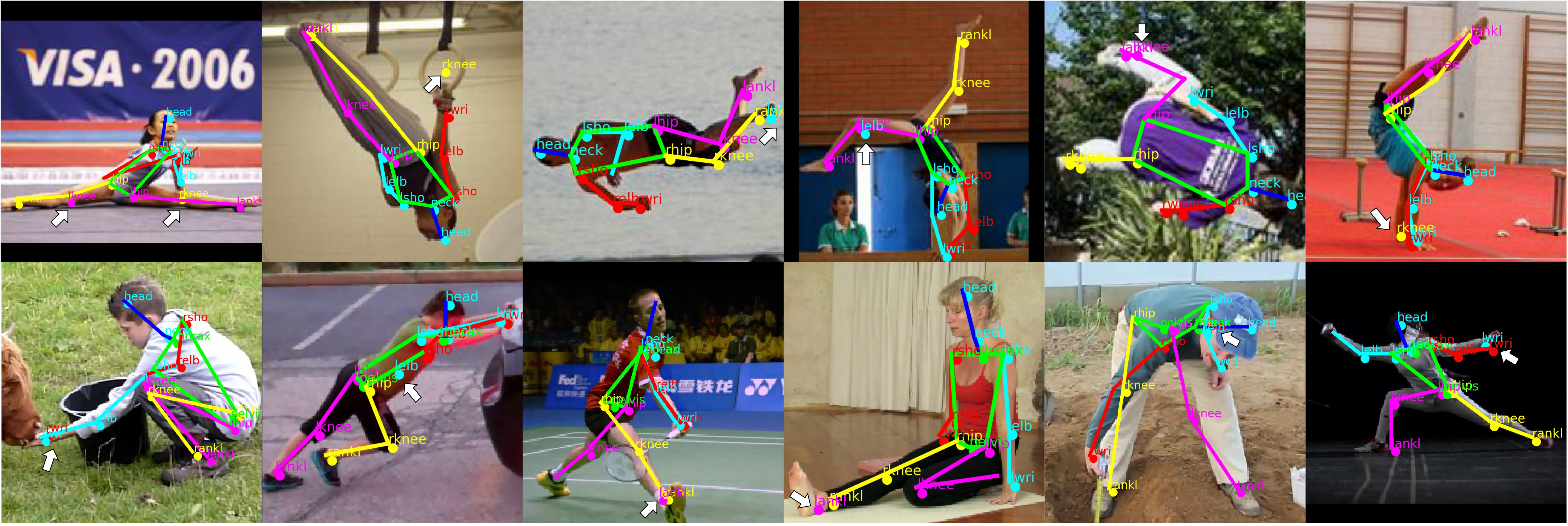}
    \caption{[Please zoom in] \textit{Pose refinement}: The skeleton represents the optimal pose configuration $Y^*$ that maximizes the likelihood, and filled circles are the the pose estimator's predictions $\hat{Y} = f(x, \Theta)$. We highlight VL4Pose's potential to identify the correct pose $Y^*$ even when $\hat{Y}$ has \underline{minor} errors (marked in arrows).}
    \label{fig:refine}
\end{figure}

We present more images depicting likelihood estimation (Fig: \ref{fig:supp_pose}) and pose refinement (Fig: \ref{fig:refine}) using VL4Pose. 

\begin{algorithm}
    \label{algo:1}
\DontPrintSemicolon
  \KwInput{Human pose estimator ($f_{\Theta}$), Auxiliary network ($g_{\theta}$), Budget ($\mathcal{B}$)}
  \KwOutput{Unlabelled images $(x^{*}_{\mathcal{U}})$ for annotation}
  \KwData{Unlabelled ($x_\mathcal{U}$) images}

  \vspace{0.25cm}
  \texttt{skeleton} = [\,  head $\rightarrow$ neck\,;\,neck $\rightarrow$ thorax $\ldots$  \,] \tcp*{length: \texttt{num\_links}}
  \vspace{0.25cm}
  
  class \texttt{Keypoint}: \\
    \hspace{0.5cm}\tcc{Default initialization for each joint}
    \hspace{0.5cm} \textcolor{purple}{\textbf{function}} \_\_init\_\_ $():$ \\
        \hspace{1cm} \textcolor{orange}{string} name \\
        \hspace{1cm} \textcolor{orange}{list} locations, peaks, children, parameters\\
    \vspace{0.25cm}
    
    \hspace{0.5cm} \tcc{DFS: Depth First Search likelihood evaluation}
    \hspace{0.5cm} \textcolor{purple}{\textbf{function}} compute\_likelihood $(\textrm{parent\_loc, link\_params}):$ \\
      \hspace{1cm} \textcolor{orange}{empty list} max\_likelihood\_per\_location \\
      \hspace{1cm} \For{i, loc in enumerate(self.locations)}
        {
        \hspace{1cm}\uIf{len (self.children) == 0}
        {
            \hspace{1.25cm} \tcc{Leaf node reached: recursion exit condition}
            \hspace{1.25cm} \textcolor{olive}{\textbf{return}} 0 \\
        }
        \hspace{1cm}\uElse
        {
            \hspace{1.15cm} \tcc{Evaluate position of self given parent location}
            \hspace{1.15cm} log\_ll = \texttt{log}\, $\mathcal{N}\,(\texttt{dist}\,(\textrm{parent\_loc\,,\,loc})\,;\,\textrm{link\_params} )$ \\
        }
        \hspace{1cm} \tcc{log prob: \texttt{log} $\hat{p}_{i}\,(y_i\,[u,v] = loc)$ where $y_i$ is heatmap $i$}
        \hspace{1cm} log\_ll += \texttt{log}\, peaks[i] \\
        \hspace{1cm} \For{child in self.children}
            {
            \hspace{1.5cm} log\_ll += child.compute\_likelihood ($\textrm{loc, parameters[i]}$) \\
            }
            \hspace{1cm} max\_likelihood\_per\_location.append(log\_ll)
        }
        \hspace{1cm} \textcolor{olive}{\textbf{return}} \texttt{max} ( max\_likelihood\_per\_location )
    
    \vspace{0.25cm}
    
  \textbf{initialize} likelihoods = empty\_array(size = $x_\mathcal{U}$.shape[0]) \\
  \vspace{0.25cm}
  \tcc{GPU parallel since each image is independent of the other}
  \For{i, $x$ in enumerate($x_\mathcal{U}$)}    
        {   
            $\bar{y} = f(x, \Theta)$ \tcp*{heatmaps of size: \texttt{num\_joints} $\times$ 64 $\times$ 64}
            $\texttt{params} = g(x, \theta)$ \tcp*{Gaussian parameters: \texttt{num\_links} $\times$ 2}
            locations, peaks = \texttt{local\_maxima} ($\bar{y}$) \\
            keypoints\_holder = dict()
            
            \vspace{0.25cm}
            \For{j, joint in enumerate(joints)}
            {
                keypoints\_holder [joint] = Keypoint (name=joint, locations[j], peaks[j])
            }
            \vspace{0.25cm}
            \For{j, link in enumerate(skeleton)}
            {
                parent = link [0] \\
                child = link [1] \\
                keypoints\_holder [parent].children.append (keypoints\_holder [child]) \\
                keypoints\_holder [parent] = \texttt{params} [j]  
                
            }
        likelihoods [i] = keypoint\_holder [`head'].compute\_likelihood()
        }
  
  \Return $x^{*}_{\mathcal{U}}$: Return samples corresponding to  \textbf{\textit{bottom - $\mathcal{B}$}} \texttt{likelihoods}
  
\caption{\textbf{\textit{VL4Key}}}
\end{algorithm}

\newpage
\bibliography{supplementary}